\documentclass[conference]{IEEEtran}
\IEEEoverridecommandlockouts
\usepackage{cite}
\usepackage{amsmath,amssymb,amsfonts}
\usepackage{algorithmic}
\usepackage{titlesec}
\usepackage{graphicx}
\usepackage{multirow}
\usepackage{textcomp}
\usepackage{bm}
\usepackage{subfigure}
\usepackage{xcolor}
\usepackage{algorithm}
\usepackage{algorithmic}

\title{SWAP: Exploiting Second-Ranked Logits for Adversarial Attacks on Time Series}

\author{
\author{
\IEEEauthorblockN{
\textbf{Chang George Dong}\IEEEauthorrefmark{1}\textsuperscript{1},
\textbf{Liangwei Nathan Zheng}\IEEEauthorrefmark{1}\textsuperscript{1},
\textbf{Weitong Chen}\IEEEauthorrefmark{2}\textsuperscript{2},
\textbf{Wei Emma Zhang}\textsuperscript{2},
\textbf{Lin Yue}\textsuperscript{3}
}
\IEEEauthorblockA{\textsuperscript{1}\textit{School of Computer Science, The University of Adelaide}\\}
\IEEEauthorblockA{\textsuperscript{2}\textit{Australian Institute for Machine Learning, The University of Adelaide}\\}
\IEEEauthorblockA{\textsuperscript{3}\textit{School of Information and Physical Sciences, The University of Newcastle Australia}\\
\{chang.dong, liangwei.zheng, weitong.chen, wei.e.zhang\}@adelaide.edu.au, \\lin.yue@newcastle.edu.au}
\thanks{\IEEEauthorrefmark{1}The first two authors contributed equally to this work.}
\thanks{\IEEEauthorrefmark{2}The third author is the corresponding author.}
}
}
\begin{document}
\maketitle
\begin{abstract}
Time series classification (TSC) has emerged as a critical task in various domains, and deep neural models have shown superior performance in TSC tasks. However, these models are vulnerable to adversarial attacks, where subtle perturbations can significantly impact the prediction results.  Existing adversarial methods often suffer from over-parameterization or random logit perturbation, hindering their effectiveness. Additionally, increasing the attack success rate (ASR) typically involves generating more noise, making the attack more easily detectable. To address these limitations, we propose SWAP, a novel attacking method for TSC models. SWAP focuses on enhancing the confidence of the second-ranked logits while minimizing the manipulation of other logits. This is achieved by minimizing the Kullback-Leibler divergence between the target logit distribution and the predictive logit distribution. Experimental results demonstrate that SWAP achieves state-of-the-art performance, with an ASR exceeding 50\% and an 18\% increase compared to existing methods.

\end{abstract}


%%%%%%%%%%%%%%%%%%%%%%%%%%Introduction%%%%%%%%%%%%%%%%%%%%%%%%%%

\section{Introduction}

Time series classification (TSC) involves assigning a class or label to a given time series data point by examining its temporal characteristics. TSC finds broad applications in various domains, ranging from complex fields like rocket science~\cite{tariq2022towards}, security~\cite{tan2017indexing} to everyday scenarios such as traffic crowded flow prediction~\cite{zhang2017deep} and power consumption monitoring~\cite{zheng2017wide}.

Deep neural models, due to their success in computer vision, have become state-of-the-art methods~\cite{krizhevsky2012imagenet}, and researchers start to adopt these methods to solve TSC problems, and they have shown significant performance improvements compared to the conventional statistical machine learning approaches~\cite{ismail2020inceptiontime}. Due to the rise of deep neural networks has also prompted investigations into their vulnerability to adversarial attacks~\cite{yuan2019adversarial}, wherein input data is manipulated with perturbations to deceive the model. Deep neural networks are often highly sensitive to minor variances in input data derived from their sharp decision boundary, which is caused by their non-linear properties. Exploiting this sensitivity, specific patterns can be discovered and can be used to harm the performance of TSC models~\cite{pialla2022smooth}. For instance, the Korean Aerospace Research Institute deployed the Convolutional LSTM with Mixtures of Probabilistic Principal Component Analyzers (CLMPPCA) method to predict anomalies on KOMPSAT-5 satellite (KARI)~\cite{tariq2019detecting}. The CLMPCA successfully predicts the anomaly for the original time series, however, with small noise perturbed by the FGSM and PGD attacks, the entire input samples to be classified as an anomaly, which can have resulted in dramatically severe consequences in the satellite safety.\cite{tariq2022towards}

\begin{figure}
\includegraphics[scale=0.5]{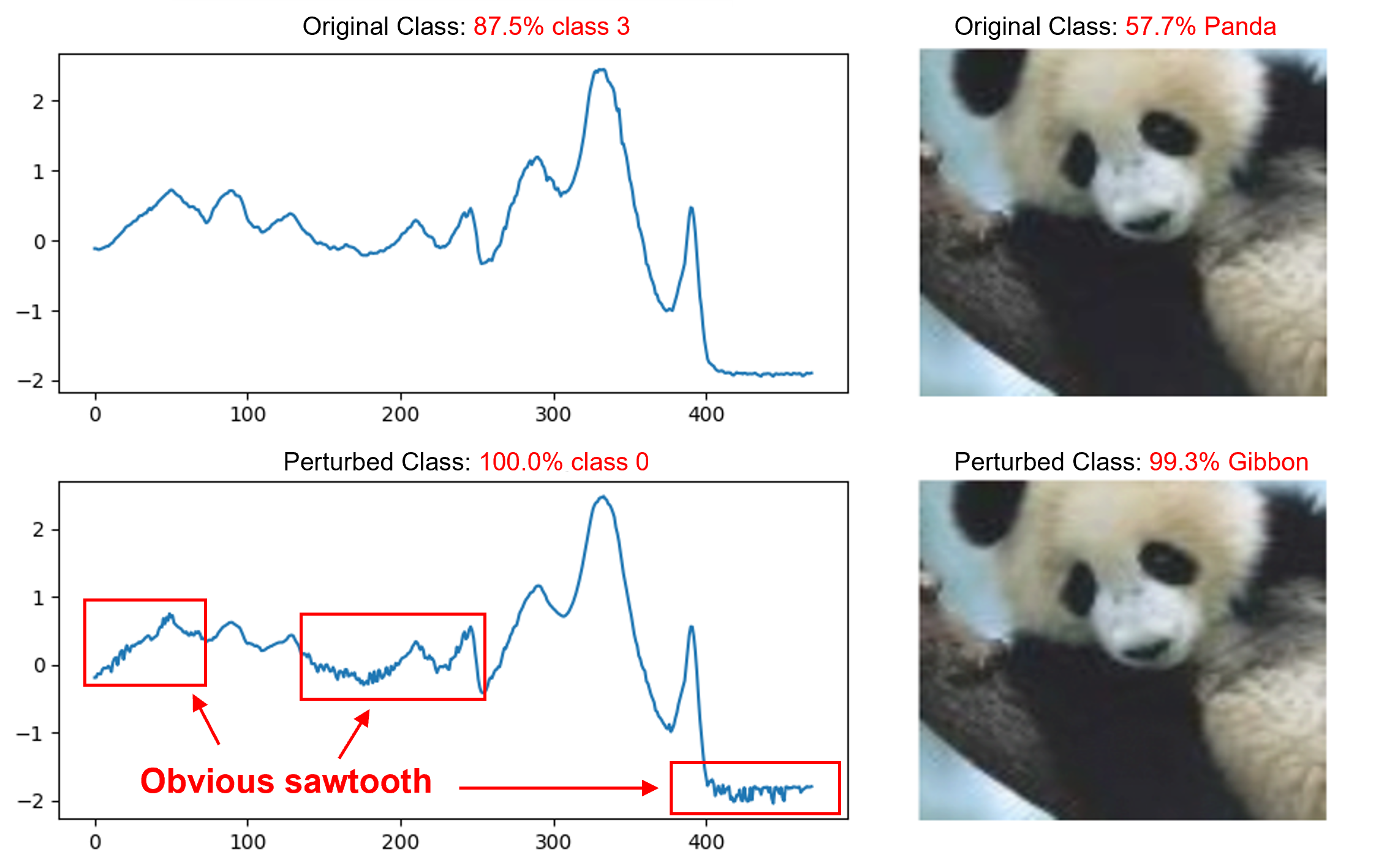}
\caption{Comparison of adversarial attacks on time series data and imaging data using the same FGSM, it is evident that adversarial attacks are generally less noticeable in images compared to time series data.}
\label{fig:attack_sample}
\end{figure}

Given the increasing abundance of time-series data and the diverse range of TSC problems encountered in the real world, it is crucial for TSC models to exhibit not only excellent performance but also sufficient robustness for accurate inference. Recently many studies explore the attacking method to TSC models, such as FGSM, Basic Interative Method (BIM), Projected Gradient Descent (PGD), and Gradient Method (GM)\cite{rathore2020untargeted,pialla2022smooth,galib2023susceptibility,fawaz2019adversarial}. The most common attacking method, FGSM and BIM and PGD which are all Gradient-Sign based methods have been proposed successfully for image recognition tasks and expanded to TSC task\cite{fawaz2019adversarial}. In the Gradient-Sign-based method, it generates perturbance by taking the gradients of the loss function with respect to the input time series data\cite{pialla2022smooth,fawaz2019adversarial}. The Gradient-Sign-based methods have demonstrated success to baffle the TSC model. But it shows poor concealment in TSC since such methods generate a significant amount of noise points that differ significantly from the original data values. While this may be tolerable for images due to the ability to conceal extreme values within pixels\cite{goodfellow2014explaining}, it becomes challenging to hide such extreme values in time series data. Consequently, the aggressive sawtooth perturbations produced by gradient-based methods can be visually distinguished as very sharp peaks within the time series. As a result, their presence becomes easily noticeable and identifiable as shown in Fig \ref{fig:attack_sample}.

To mitigate the sawtooth pattern in time series data caused by the adversarial attack, the GM with regularization has been proposed to enhance the stealthiness of attack noise. By employing $\mathcal{L}^{2}$ norm regularization to constrain the noise level, GM reduces nearly half the level of noise by comparing it to the Basic Iterative Methods~\cite{pialla2022smooth}. Building upon the success, Pialla et al. proposed Smoothed Gradient-based method (SGM) which incorporates an additional $\mathcal{L}^{1}$ regularization terms to further minimize the difference between the successive attack time series data to the original data. As a result, SGM generates smoother noise compared to the former methods. However, a trade-off involved compromising the attack success rate (ASR).

Despite the promising results of existing approaches in adversarial attacks, they often suffer from issues such as over-parameterization and random logit perturbation. The conventional method involves a random selection of logits from the output, excluding the predicted class, and subsequently increasing their confidence level to confuse the TSC model. However, this random selection strategy may not be optimal, especially when dealing with logits that possess low confidence. Consequently, this approach can lead to a low success rate in the attack. Furthermore, when there is a substantial gap in logits before the softmax activation between the prediction and randomly selected logits, training the model to improve confidence in the selected logits becomes challenging. Increasing the attack success rate usually requires generating more noise and applying it to the original input. However, this can also make the attack easily identifiable. Random selection of logits, while effective in altering the least confident class, can introduce a significant amount of noise that might raise suspicions and expose the attack.

To address the aforementioned issues, we introduce a novel adversarial attack framework called SWAP in this paper. Unlike the aforementioned approaches, the SWAP framework specifically aims to enhance the confidence of the second-largest logits while minimizing manipulation on the other logits. The key aspect of our framework is the targeted swapping of ranks between the prediction logits and the target logits, achieved by minimizing the Kullback-Leibler divergence ($D_{KL}$) between the distribution of target logits and the perturbed logits. This targeted swapping approach focuses on perturbing the logits that are most relevant to the target class, leading to improved attack effectiveness. 

The contributions of this work can be summarized as follows:

\begin{itemize}
    \item Novel adversarial attack framework: A novel adversarial attack framework namely, SWAP, is proposed, different from the existing method, SWAP focuses on enhancing the confidence of the second largest logits while minimizing manipulation of other logits. This targeted approach increases attack effectiveness by perturbing the most relevant logits to the target class.
    \item Rank swapping for improved attack success: SWAP leverages the concept of position swapping, aiming to exchange the position of the prediction logits with the target logits. By minimizing the Kullback-Leibler divergence %($D_{KL}$) 
    between the distributions of the target logits and the perturbed logits, we achieve a more effective perturbation. % strategy.
    \item Consideration of logits' importance: We address the issue of randomly selecting logits for perturbation by focusing on enhancing the confidence of specific logits. By prioritizing the manipulation of the second largest logits, we improve the attack's success rate while reducing the potential for easy identification.
    \item Experimental evaluation: Intensive experiments are conducted to evaluate the effectiveness of the SWAP framework and demonstrate that SWAP achieves higher attack success rates while minimizing the amount of generated noise.
\end{itemize}

\begin{figure*}[h]
    \centering
    \includegraphics[scale=0.55]{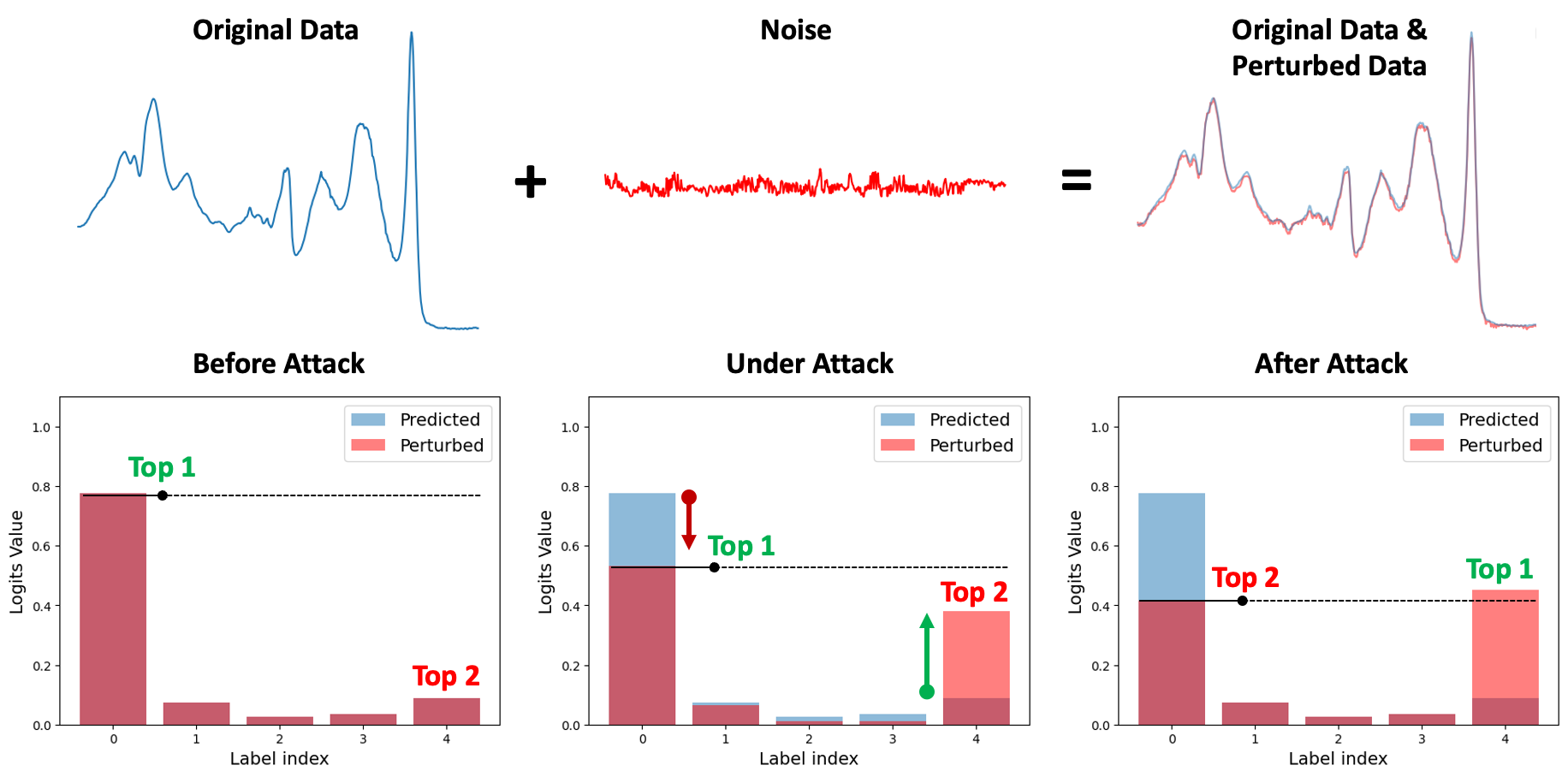}
    \caption{Schematic diagram of the attacking process, example data from Beef dataset, attacked by SWAP method}
    \label{fig:attack_scheme}
\end{figure*}

\section{Relative Works}

Deep neural models  have achieved significant success across tasks in the computer vision domain, such as image classification, object detection, and semantic segmentation. However, these models are not without limitations. Research by Szegedy et al. \cite{eykholt2018robust,goodfellow2014explaining,madry2017towards} has shown that even imperceptible levels of noise can have a drastic impact on the output of these models. This phenomenon is not limited to specific network complexities; it is pervasive across deep neural models. While deep neural models have demonstrated remarkable performance in computer vision tasks, it is crucial to address their susceptibility to adversarial attacks\cite{akhtar2021advances}. Ongoing research endeavours continue to explore techniques and strategies to enhance the resilience of deep neural models against such attacks, aiming to ensure their reliable and safe deployment in practical scenarios.

\subsection{Adversarial Attack}
Adversarial attacks can be categorized into two main types: Black-Box attacks and White-Box attacks\cite{ebrahimi2018hotflip,jiang2019black}.
In the White-Box attack, the attacker has complete knowledge and access to all information about the targeted model and dataset. While the Black-Box attack, the attack has limited or no access to specific information about the target model. Most of the research in the field has focused on White-Box attacks targeting image classification models\cite{yuan2019adversarial}. For example, Goodfellow et al. develop the FGSM\cite{wang2021adversarial}, Then, Kurakrin et al. \cite{pialla2022smooth} provide an alternative to expensive optimization techniques Basic Iterative Method (BIM). BIM seeks to find perturbations that maximize the model's loss on a specific input while keeping the noise amplitude below a certain constraint called epsilon ($\epsilon$). Notably, even printed photos have been shown to effectively fool the model \cite{tariq2022towards, wang2021adversarial, eykholt2018robust}. Another powerful method is Projected Gradient Descent (PGD)\cite{deng2020universal}, which is similar to BIM in iteratively updating the noise but choosing a random initial point to achieve better performance. PGD has demonstrated effectiveness in generating adversarial examples in various domains. These methods highlight the importance of understanding and mitigating the vulnerabilities of deep learning models to adversarial attacks. Researchers have explored a range of attack techniques and defence mechanisms to enhance the robustness and security of deep neural networks in the face of adversarial examples.

\subsection{Time Series Adversarial Attack}
Given the successful application of adversarial attacks in image classification, researchers have also explored the extension of these methods to Time Series Classification models. Fawaz et al. focus on univariate datasets from the UCR repository and adapted existing adversarial attack methods for TSC \cite{fawaz2019adversarial}. Gautier Pialla et al.\cite{pialla2022smooth} highlight the ability to achieve complete changes in the output by maximizing the Kullback-Leibler divergence between the predicted and perturbed logits. The KL divergence measures the differences between the proposed distribution and the target distribution. By introducing regularization techniques, they are able to obtain smoother disturbances. However, it did not maximize the KL divergence directly, instead by minimizing the cross entropy between the target logits, which is randomly selected a label as a one-hot distribution,  and perturbed logits, turning this problem into a one-hot cross-entropy minimizing the problem. This random selection can seriously lower the ASR and increase the noise level, especially when there is a substantial gap in logits before the softmax activation between the prediction and randomly selected logit. 

In this paper, we only focus on the White-Box Attack to reveal the robustness of a specific model in different datasets. 

% \subsection{Adversarial attack}
% what is Adversarial attack in general?

% What is Black and white box?

% \subsection{Time Series attack}

% What is TSC adversarial attack

% Some general approach to adversarial attack in TSC 

% limitation(briefly introduce)

% \subsection{KL-divergence}
% KL-divergence and CE differences, graidient Sign, the problem of CE. Why KL?

%%%%%%%%%%%%%%%%%%%%%%%%%% Preliminary %%%%%%%%%%%%%%%%%%%%%%%%%

\section{Preliminary}
To facilitate reproducibility and enhance understanding of our proposed method, we provide key definitions and concepts. Our method is a White Box Adversarial Attack, where the model architecture and dataset are known. In our attack strategy, we aim to perturb a Time Series Data $X$ in order to divert the model's prediction from the original class $Y_i$. The attack is considered successful if the perturbed class $Y_m$ replaces the original prediction. We define the following:

\begin{itemize}
    \item \textbf{Definition 1} A Univariate Time Series $X = [x_1, x_2,...,x_T]$  with all $x_i \in \mathbb{R}^M$ and $T$ denoting its length.
    \item \textbf{Definition 2} $D = \{(X_1,Y_1),...,(X_N,Y_N)\}$ is the the Dataset, with N samples, and $Y_i$ is an one-hot label vector.
    \item \textbf{Definition 3} $f(x)$ denotes the output logits of TSC model given an input $x$. $f(x')$ refers to the logits after being attacked, and $Y^t$ is the designed target logits (All the logits are $softmax$ logits).
    \item \textbf{Definitiion 4} Successful Adversarial attack aims to train a $x' = x + r$, where $r$ is the noise and $x'$ refers to the perturbed time series, leading $arg\ max{\ f(x')} \ne arg\ max{\ f(x)}$. And the $x'$ and $x$ should keep close to each other to avoid being perceived visually.
    \item \textbf{Definition 5} Consider two probability distributions $f(X')$ and $f(X)$. Here, $f(X)$ represents the ground truth probability distribution, while $f(X')$ represents an approximation of the ground truth sample. The Kullback-Leibler Divergence\cite{cui2023decoupled}, denoted as $D_{KL}$, is a measure of the difference between these two probability distributions, $f(X)$ and $f(X')$. Higher $D_{KL}$ indicates a closer similarity of two probability distributions. For discrete probability distributions, $D_{KL}$ is defined as follows:
    \begin{equation}
    D_{KL}(f(X),f(X'))\equiv \sum^{c}f(X)log\frac{f(X)}{f(X')} \label{eq: D_KL}
    \end{equation}
    where $c$ is the number of classes for time series data $X$.
    \item \textbf{Definition 6} $\mathcal{L}^{2}$ Norm Regularization is one of the most important techniques to constrain the complexity of the model resulting in smoother fitting and higher generalization ability. Thus, given an noise $r$ and a regularization parameter $\alpha$, the regularization is defined as follow:
    \begin{equation}
    \mathcal{L}^{2} = \alpha \cdot ||r||_2 \label{eq:L2}
    \end{equation}
    \begin{equation}
    ||r||_2 = \sum_{i=0}^{l}|r_i| \label{eq:r_2}
    \end{equation}
\end{itemize}

The Gradient Method (GM) with $\mathcal{L}^{2}$ norm, developed by Pialla et al. \cite{pialla2022smooth}, effectively maximizes the difference between the original logit distribution $f(X)$ and the perturbed logit distribution $f(X')$, by maximizing the $D_{KL}$. The hyperparameter $\mu$ controls the penalty for misclassification. The $\mathcal{L}^{2}$ norm in Equation 7 constrains the noise amplitude.

To further mitigate the sawtooth pattern of the noise, $\mathcal{L}^{1}$ norm term, which constraints the distance between all the two adjacent times, $||r_t,r_{t+1}||_1$, can be introduced to the GM($\mathcal{L}^{2}$) framework, which is called Smooth Gradient Method(SGM). Eq.\ref{eq:SGM} denotes the application of $\mathcal{L}^{1}$ in GM($\mathcal{L}^{2}$) framework.

The GM method randomly selects logits and increases their confidence to suppress the original prediction. It then minimizes the difference between the target and perturbed logit distributions. The randomly selected logits are adjusted to be as close to 1 as possible, suppressing the logits of the original prediction.

\begin{equation}
X' = X + r\label{eq}
\end{equation}
\begin{equation}
D_{KL}(f(X),f(X'))\equiv \sum^{c}f(X)log\frac{f(X)}{f(X')} \label{eq:D_KL_GM}
\end{equation}

\begin{equation}
GM = - D_{KL}(f(X),f(X'))\label{eq:GM}
\end{equation}

\begin{equation}
GM(\mathcal{L}^{2}) = \mu\cdot GM+ \alpha\cdot ||r||_2\label{eq:GM_L2}
\end{equation}
\begin{equation}
SGM = GM(\mathcal{L}^{2}) + \gamma\cdot \sum_{i=0}^T||r_i, r_{i+1}||_1\label{eq:SGM} 
\end{equation}

Although the GM method does not operate on $D_{KL}$ as shown in Eq. \ref{eq:D_KL_GM}, our method is inspired by GM to minimize the $D_{KL}$ between target distribution and perturbed distribution.

% logits selection

% FGSM(briefly, move to experiment)
% BIM(briefly, move to experiment)
% GM
% GM(L2)
% SGM

\section{Proposed Method}
%%%%%%%%%%%%%%%%%%%%%%%%%% Proposed Method %%%%%%%%%%%%%%%%%%%

To successfully execute an adversarial attack on a model, the perturbation introduced should result in a change in the model's output, specifically altering the highest-ranking prediction. The goal is to manipulate the model in such a way that the original top-ranked prediction is demoted to a lower rank, even if it is just to the second rank. This change in ranking serves as an indication of a successful attack, as it signifies a significant alteration in the model's output.

Furthermore, through our observations, we have noticed a positive relationship between the magnitude of the noise perturbation, measured by the Euclidean distance, and the KL divergence between the predicted and perturbed logits. This suggests that a greater disparity in the logit distribution is associated with a higher level of noise perturbation. Consequently, reducing the dissimilarity between the original and perturbed logits can yield significant benefits in terms of reducing the overall noise level introduced during the attack. By minimizing the divergence between the target logits and the perturbed logits, we aim to mitigate the noise level and enhance the stealthiness of the attack. This approach allows us to achieve a more effective perturbation strategy by focusing on reducing the discrepancy between the distributions of logits. By doing so, we can generate adversarial samples that have a lower noise magnitude while still inducing a significant change in the model's output.

\begin{figure}[h]
    \centering
    \includegraphics[scale=0.5]{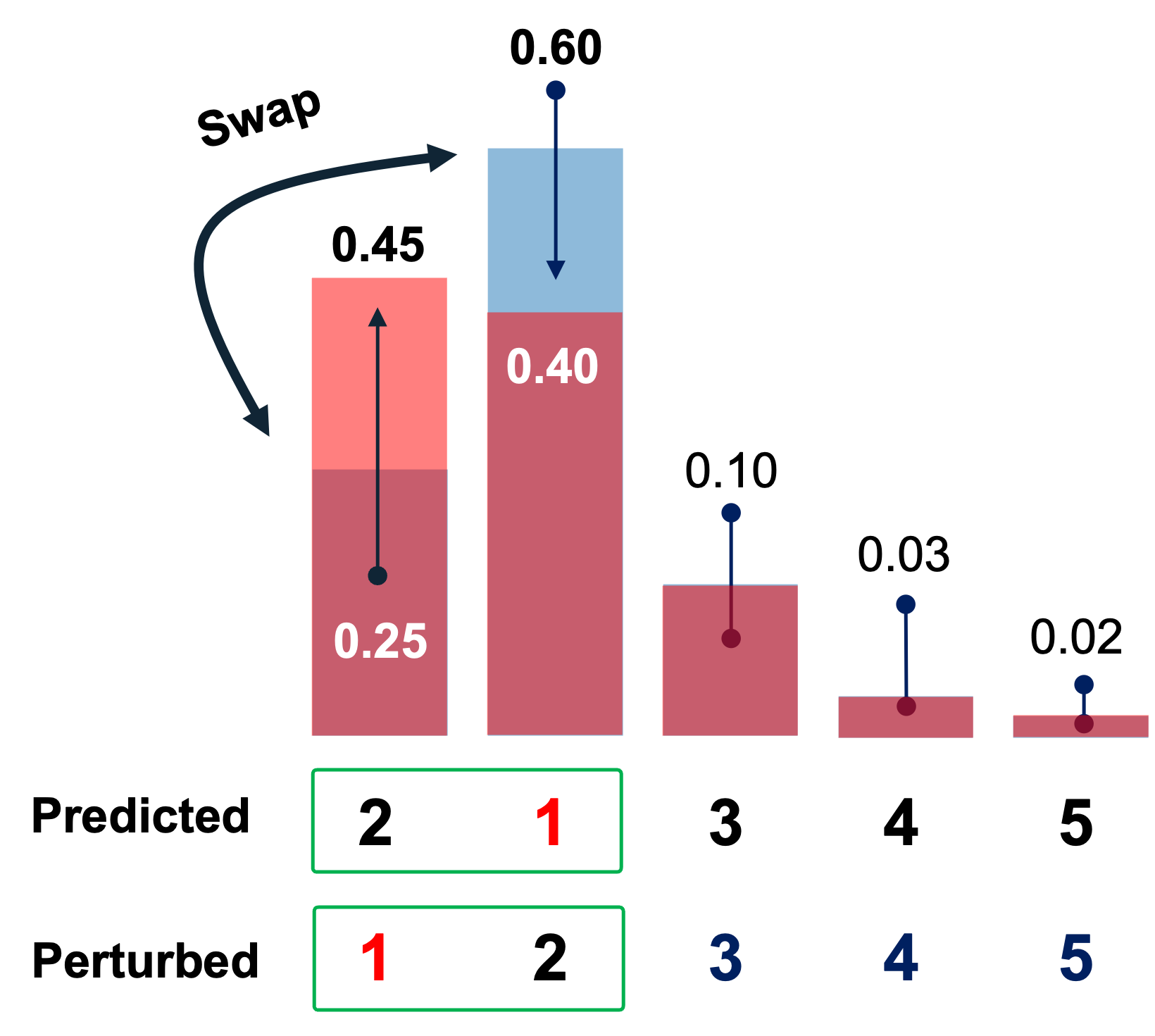}
    \caption{Illustration of Swap Method}
    \label{fig:Illustration of Swap}
\end{figure}

However, for a successful adversarial attack, the minimal change in logits that is required is simply a rearrangement of the top two logits: the second highest logit just needs to slightly surpass the predicted one, while all other logits remain unchanged as shown in Fig. \ref{fig:Illustration of Swap}. Therefore, we have designed a target logit distribution. By minimizing the KL-divergence between the logits after the attack and this target logit distribution, we can ensure that the resulting logit distribution is as close as possible to the original distribution, while still achieving a successful attack. Thus, we proposed the SWAP method, to preserve the features of time series data to the utmost extent while the model is successfully attacked.

\begin{equation}
f(x_1)^t = \gamma(f(x_1)^o + f(x_2)^o) \label{eq:logits_original}
\end{equation}

\begin{equation}
f(x_2)^t = (1 - \gamma)(f(x_1)^o+f(x_2)^o) \label{eq:logits_target}
\end{equation}

Here, we selected the second-ranked logits of predictive distribution $f(x_2)^o$ as our distractive logits, and swap the rank of our distractive logits and the prediction $f(x_1)^o$ by the balance factor $\gamma$ as shown in Eq. \ref{eq:logits_original} and Eq. \ref{eq:logits_target}

\begin{equation}
D_{KL}(f(X)^{t}|f(X') = \sum^{c}_{i=0} f(x_i)^tlog\frac{f(x_i)^t}{f(x_i)}\label{eq:D_KL_us}
\end{equation}
   
\begin{equation}
SWAP = D_{KL}(f(X)^{t}|f(X'))\label{eq:SWAP}
\end{equation}

\begin{equation}
SWAP(\mathcal{L}^{2}) =  SWAP + \alpha||r||_2 \label{eq:SWAP_L2}
\end{equation}

SWAP takes a different approach compared to existing methods by considering the strong relationships between logits and the features of time series data. Instead of randomly selecting logits and minimizing the differences between the perturbed logit distribution and the target logit distribution, SWAP focuses on the second-ranked logits of the predictive distribution $f(x_2)^o$ as the distractive logits. The rank of the distractive logits is then swapped with the prediction $f(x_1)^o$ using a balance factor $\gamma$ as shown in Eq. \ref{eq:logits_original}. The typical value of $\gamma$ is 0.48. By performing this rank swapping, SWAP aims to achieve a more targeted and effective perturbation strategy.

To further align the distributions and reduce discrepancies, SWAP minimizes the Kullback-Leibler divergence ($D_{KL}$) between the target distribution $f(X_i)^t$ and the perturbed distribution $f(X_i^{'})$. This minimization step helps to bring the distributions closer together and reduce the differences between them. As a result, the features of the original time series data are preserved to a significant extent while still achieving a successful attack. 

As a result, the features of the original time series data will be preserved at the utmost extend while the attack is successfully achieved as shown in Fig. \ref{fig:attack_scheme}.
% \begin{figure}[h]
%     \centering
%     \includegraphics[scale=0.5]{Figure/KL_dist.png}
%     \caption{Relationship between $D_{KL}$ of predicted and perturbed logits and Noise level}
%     \label{fig:Relationship of DKL and distance}
% \end{figure}

% \begin{figure}[h]
%     \centering
%     \includegraphics[scale=0.4]{Figure/Procedure.png}
%     \caption{}
%     \label{fig:Relationship of DKL and distance}
% \end{figure}

\begin{algorithm}
\caption{SWAP Adversarial Attack}
\begin{algorithmic}[1]
\REQUIRE Original time Series $X$, pre-trained model $f$, parameter $\gamma$
\ENSURE Perturbed time series $X'$
\STATE Initialize: $r$
\STATE Initialize: $X' = X + r$
\STATE Design target logits: $Y^t$ (SWAP)
\FOR{$i = 1$ to $1000$}
    \STATE ${\hat{Y'}} = f(X')$
    \STATE $r = r - \beta \cdot \nabla_r D_{KL} (Y^t, \hat{Y}')$
    \STATE $X' = X + r$
\ENDFOR
\STATE $X' = \min\{X + \epsilon, \max\{X - \epsilon, X'\}\}$
\end{algorithmic}
\end{algorithm}

%%%%%%%%%%%%%%%%%%%%%%%%%%Experiment%%%%%%%%%%%%%%%%%%%%%%%%%%

\section{Experiment}

\subsection{Dataset}\label{AA}
In our experimental evaluation, we compared the proposed method with a baseline approach using the UCR Archive-2018 dataset as the target for our adversarial attacks. The UCR Archive-2018 dataset\cite{dau2019ucr} comprises 128 diverse time series types from various domains, including healthcare, agriculture, finance, engineering, and more \cite{dau2019ucr}. For our analysis, each dataset in the UCR Archive-2018 is divided into training and testing sets.

To conduct the attacks, we employed the InceptionTime\cite{ismail2020inceptiontime} time series classifier, which is a widely used model known for its robustness and generalization capabilities in time series classification tasks. InceptionTime utilized a ResNet deep learning architecture\cite{he2016deep} specifically designed for time series analysis. For training the InceptionTime model, we utilized the best parameters and settings as provided by \cite{ismail2020inceptiontime}, ensuring a reliable and consistent baseline for our experiments.

\subsection{Environment and Setting}\label{AA}
All the codes including the proposed method and baseline method of our experiments were available on GitHub\footnote{Due to the triple-blind review process and to ensure anonymity, the link to the repository will be made available once the publication is finalised}. Our experiments were conducted on a server equipped with 2 Nvidia RTX 4090 GPUs, 64 GB RAM, and an AMD EPYC 7320 processor.

For our SWAP method, we used the same parameter settings as the GM and other existing methods to ensure fair and consistent comparisons. Hence, we applied 1000 iterations for the attack process, with a noise clipping value set to 0.1. Additionally, we introduced a parameter, $\gamma$, with a value of 0.48 to control the ratio between the logits of the original prediction and the logits of the second largest class. The scaling coefficient, denoted as $\alpha$, for the L2 regularization term was set to 0.01 as the default value.

\subsection{Evaluation Metrics}\label{AA}
\begin{itemize}
    \item \textbf{Attack Succeeded Rate(ASR)}: 
    ASR\cite{karim2020adversarial} is defined as the rate of successful attacks after applying the adversarial noise. It represents the success rate of the noise in confusing the targeted model, resulting in incorrect predictions.
        \begin{equation}
        ASR=\frac{|X_t|}{|X_f|+|X_t|}
        \end{equation}
        where $|X_f|$ is the number of unsuccessful attacked samples and $|X_t|$ is the number of successful attacked samples.

    \item \textbf{Average Distance}: Average Distance\cite{karim2020adversarial} is defined as the mean difference in amplitude at each time step between the attacked series and the original series, which are all obtained in the successfully attacked samples.
    \begin{equation}
        Average Distance =||X' - X||_2
    \end{equation}
    where ${X^{'}}$ is the modified time series data and ${X}$ is the original data. The higher value indicates that a more violent noise is applied.
\end{itemize}

\subsection{Implementation}\label{AA}
We implemented our SWAP method along with other commonly used adversarial attack methods, namely FGSM, BIM, GM, and SGM, on the InceptionTime model. We measured the Attack Success Rate and Average Distance as performance metrics to compare the effectiveness of these different attack methods. To account for result variations, we conducted multiple tests using different random seeds. Additionally, we investigated the relationship between the amount of noise introduced and the resulting distance metric. Moreover, we compared the performance of our proposed SWAP method with various baseline models on the same dataset and TSC model. This comprehensive evaluation allowed us to assess the superiority of our method against existing approaches in terms of both ASR and the distance metric. \textbf{SWAP and SWAP($\mathcal{L}^{2}$)}, the proposed method and its variation to suppress the prediction by $\gamma$, and minimize the $D_{KL}$ between the target distribution and perturbed distribution, we choose our balance factor $\gamma$ as 0.48 indicating one-step alteration of predictive class and $\mathcal{L}^{2}$ regularization parameter as 0.1.

\begin{figure*}[h]
    \centering
    \includegraphics[scale=0.43]{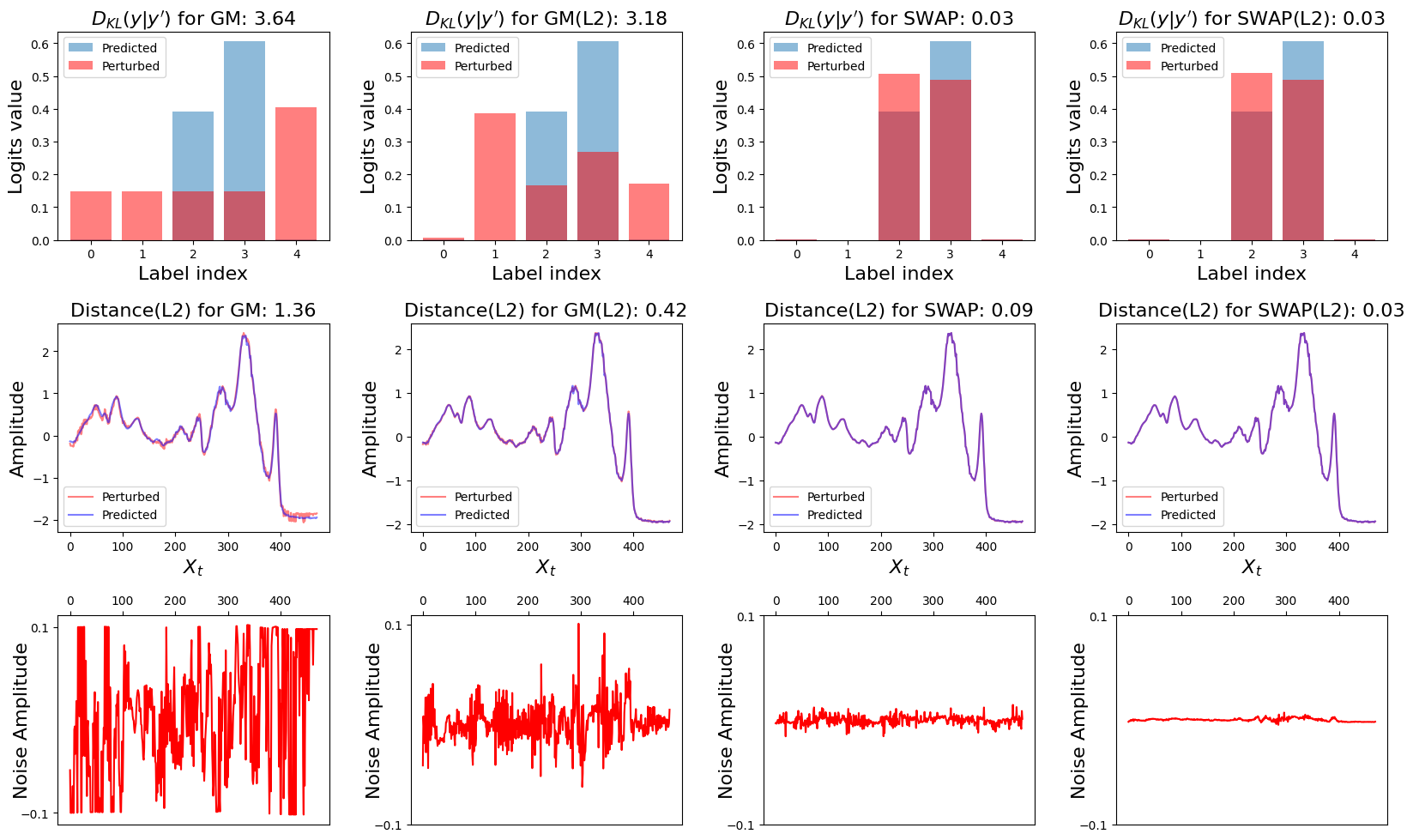}
    \caption{Scheme of adversarial attack, example data from Beef dataset, attacked by GM and SWAP}
    \label{fig:visualynalusis}
\end{figure*}
\begin{figure*}[htbp]
    \centering
    \subfigure{
        \includegraphics[scale=0.28]{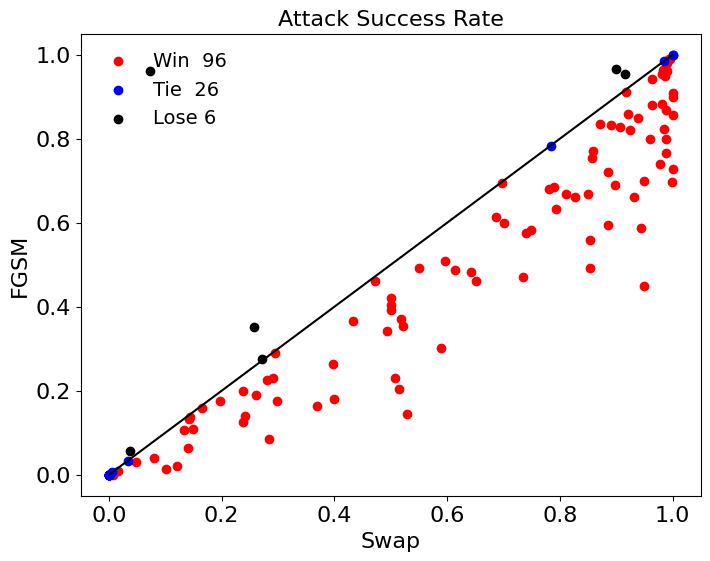}
        \label{fig:subfig_a}
    }
    \subfigure{
        \includegraphics[scale=0.28]{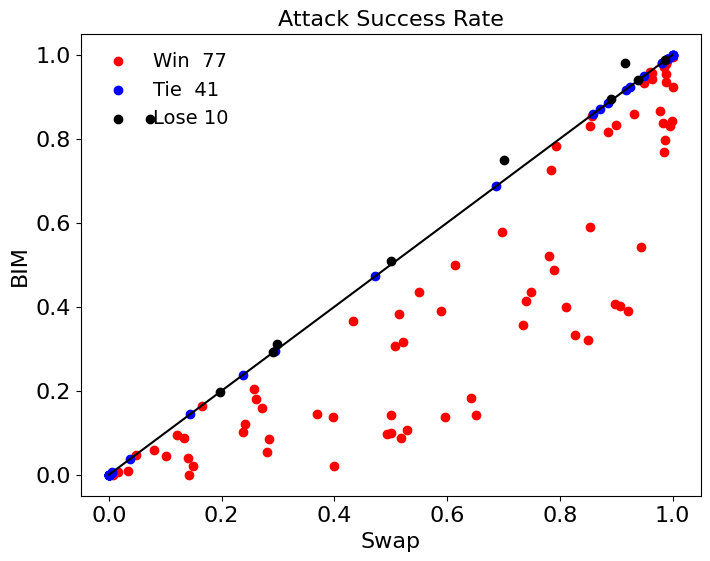}
        \label{fig:subfig_b}
    }
    \subfigure{
        \includegraphics[scale=0.28]{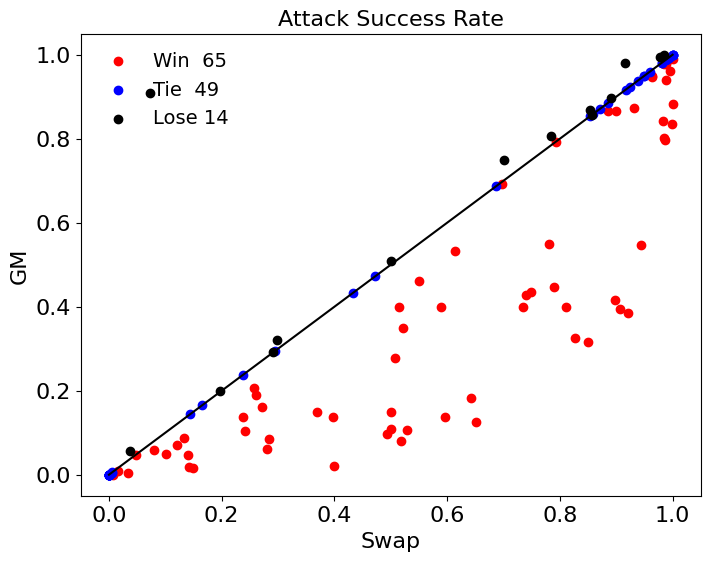}
        \label{fig:subfig_c}
    }
    \subfigure{
        \includegraphics[scale=0.28]{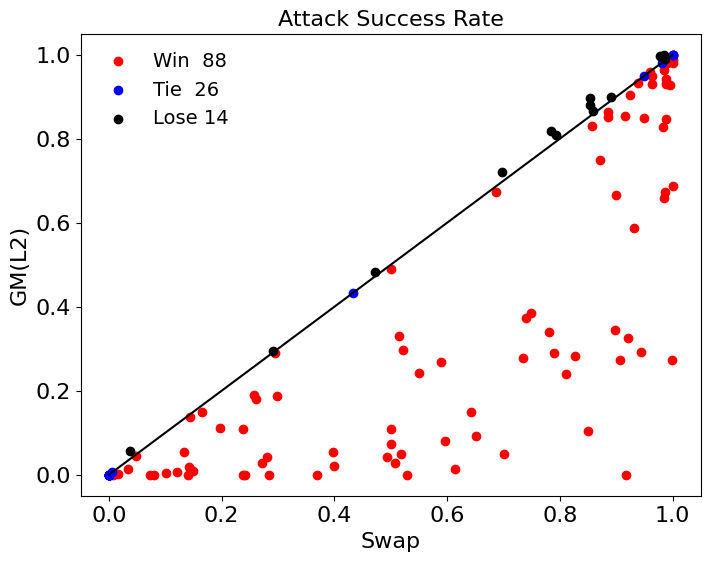}
        \label{fig:subfig_d}
    }
    \subfigure{
        \includegraphics[scale=0.28]{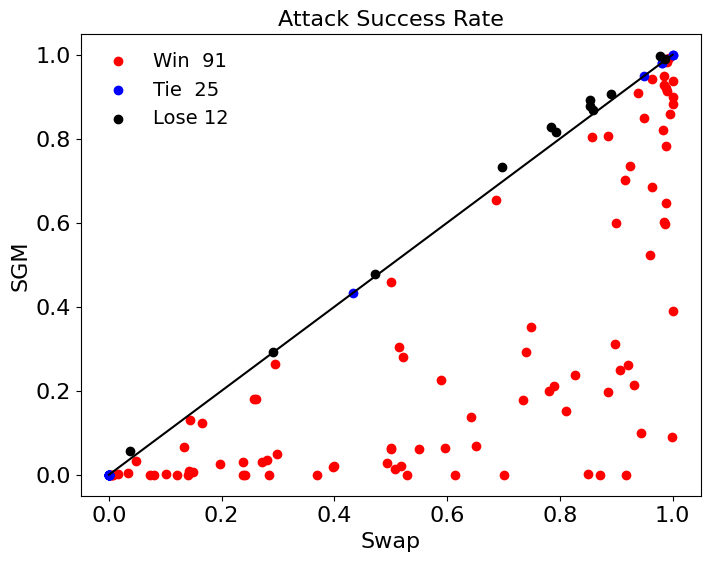}
        \label{fig:subfig_e}
    }
    \subfigure{
        \includegraphics[scale=0.28]{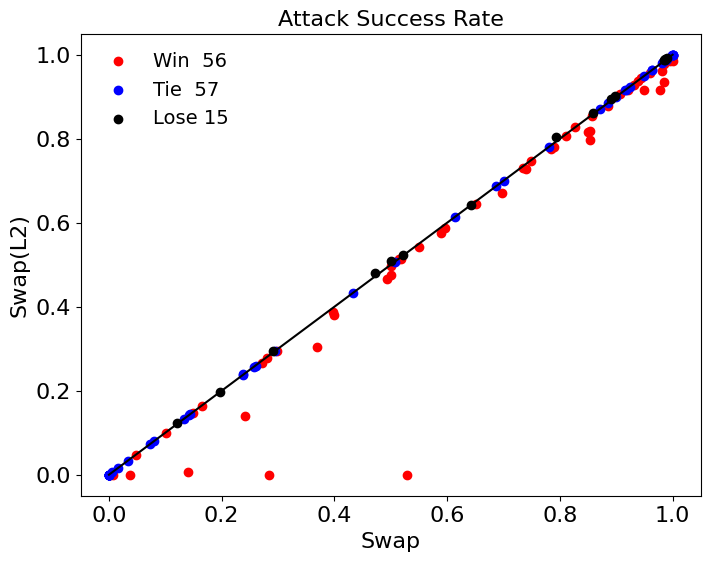}
        \label{fig:subfig_f}
    }    
    \caption{ASR comparison of SWAP vs FGSM, BIM, GM, GM($\mathcal{L}^{2}$), SGM and SWAP($\mathcal{L}^{2}$) algorithm.}
    \label{fig:ASR comparison}
\end{figure*}

\begin{figure*}[htbp]
    \centering
    \subfigure{
        \includegraphics[scale=0.28]{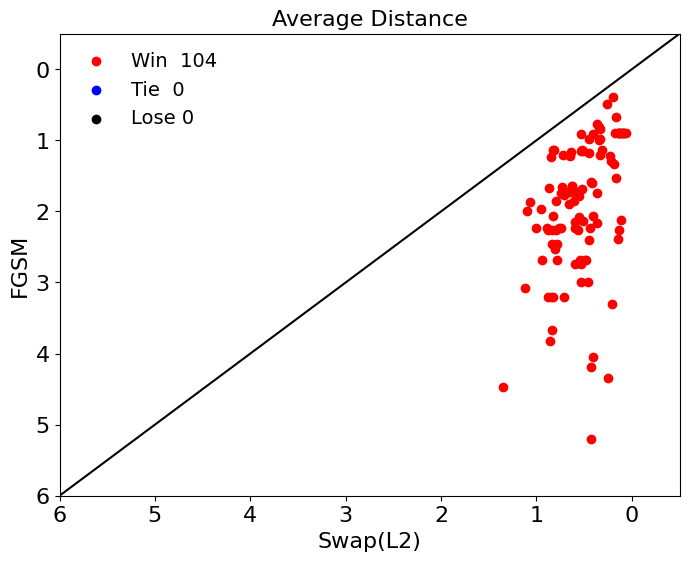}
        \label{fig:subfig_a}
    }
    \subfigure{
        \includegraphics[scale=0.28]{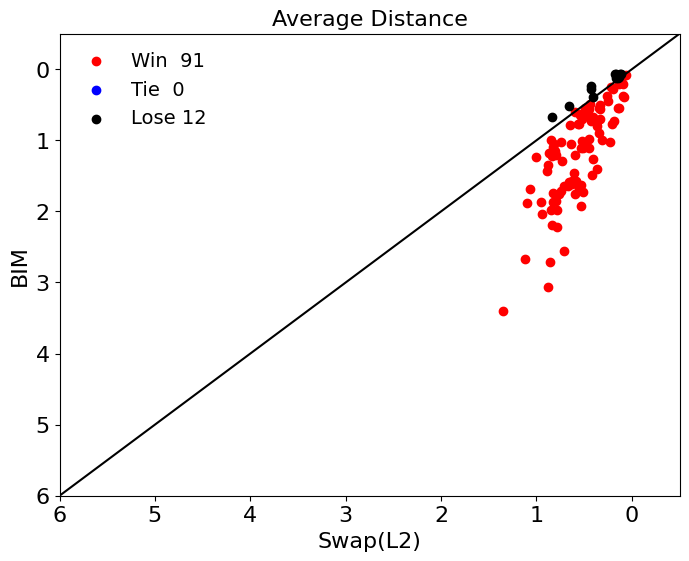}
        \label{fig:subfig_b}
    }
    \subfigure{
        \includegraphics[scale=0.28]{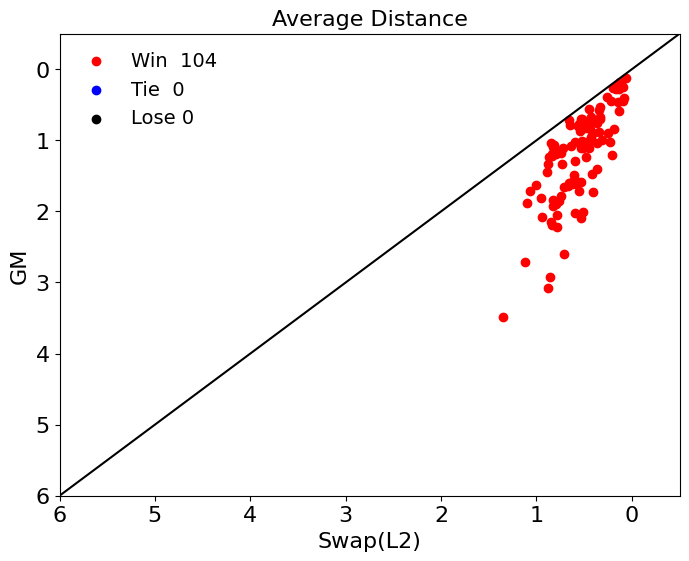}
        \label{fig:subfig_c}
    }
    \subfigure{
        \includegraphics[scale=0.28]{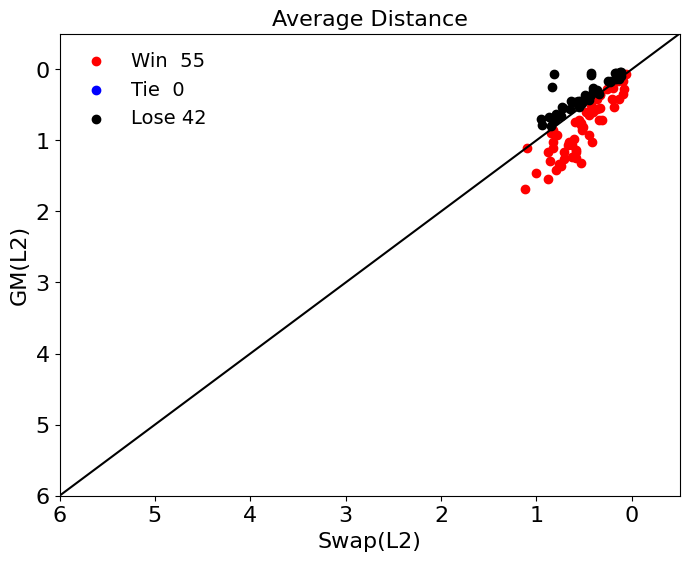}
        \label{fig:subfig_d}
    }
    \subfigure{
        \includegraphics[scale=0.28]{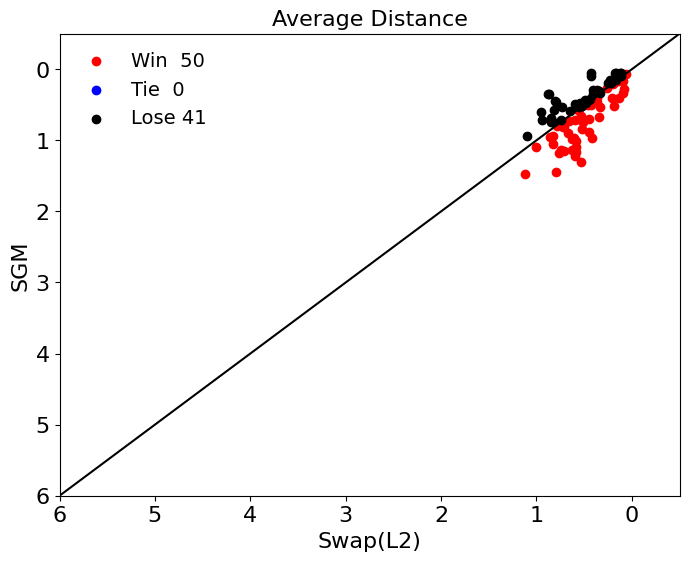}
        \label{fig:subfig_e}
    }
    \subfigure{
        \includegraphics[scale=0.28]{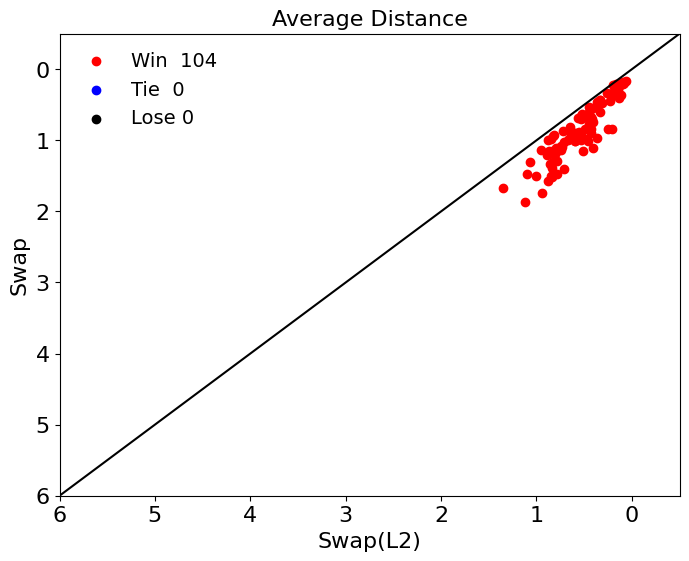}
        \label{fig:subfig_f}
    }    
    \caption{Distance comparison of SWAP vs FGSM, BIM, GM, GM($\mathcal{L}^{2}$), SGM and SWAP($\mathcal{L}^{2}$) algorithm.}
    \label{fig: Distance Comparison}
\end{figure*}

\textbf{Baseline Methods}

\begin{itemize}

    \item \textbf{FGSM}\cite{fawaz2019adversarial} takes the opposite direction of the gradient with a scaling factor $\beta$ to generate a noise. The scaling factor $\beta = 0.1$  is selected as suggested to generate the imperceptible noise.

    \item \textbf{BIM} is the iterative version of FGSM to evolve the noise generated from gradient and introduce a clip bound $\epsilon$ to constrain the amplitude of the noise. The scaling factor $\beta = 0.0005$ for 1000 iterations and the clip bound $\epsilon = 0.1$ is adopted as advised.

    \item \textbf{GM}\cite{pialla2022smooth} method randomly selects a logit except the prediction to enlarge the confidence and maximizes the $D_{KL}$ between target distribution and perturbed distribution. The scaling factor $\beta = 0.0005$ and clip bound $\epsilon = 0.1$ are used for the best performance.

    \item \textbf{GM($\mathcal{L}^{2}$)} is a variant of GM method, $\mathcal{L}^{2}$ regularization is employed to produce smoother noise by constraining the amplitude of the noise. The $\mathcal{L}^{2}$ regularization factor $\alpha = 1$ is confined as suggested. 

    \item \textbf{SGM} is a variant method that attached a $\mathcal{L}^{1}$ regularization to GM($\mathcal{L}^{2}$), constrained the difference of two noise for the adjacent time step producing smoother attack. $\mathcal{L}^{1} = 0.1$ is chosen by suggested.

\end{itemize}

% \begin{table}[!ht]
%     \centering
%     \caption{Metrics comparison}
%     \renewcommand{\arraystretch}{1.5}
%     \begin{tabular}{c|c|c|c}
%         \hline
%          \multirow{2}{*}{\textbf{Method}}  & \multirow{2}{*}{\textbf{ASR}}  & \multirow{2}{*}{\shortstack{\textbf{Average Distance}}}  &  \multirow{2}{*}{\textbf{Average Distance}} \\
%          & & & \\
%          \hline
%          SGM        & $0.3238$            &  0.5735  & 0.5087    \\
%          GM($\mathcal{L}^{2}$)     & $0.3736$            &  0.6312  &  0.5613       \\
%          BIM        & $0.4354$            &  1.1280  &  1.2620     \\
%          GM         & $0.4453$            &  1.2505  &  1.3379       \\
%          FGSM       & $0.4517$            &  1.9843  &  1.8587    \\
% \multirow{2}{*}{\textbf{SWAP}} & \multirow{2}{*}{\shortstack{\textbf{0.5339}\\$\pm$\textbf{9.93e-4}}} & \multirow{2}{*}{\shortstack{\textbf{0.7661}\\$\pm$\textbf{7.60e-2}}} &  \multirow{2}{*}{\shortstack{\textbf{1.1324}\\$\pm$\textbf{4.76e-2}}}   \\
%           & & & \\
% \multirow{2}{*}{\textbf{SWAP($\mathcal{L}^{2}$)}} & \multirow{2}{*}{\shortstack{\textbf{0.5231}\\$\pm$\textbf{7.64e-4}}} & \multirow{2}{*}{\shortstack{\textbf{0.5374}\\$\pm$\textbf{3.74e-3}}} & \multirow{2}{*}{\shortstack{\textbf{0.7469}\\$\pm$\textbf{2.52e-3}}} \\
%           & & & \\
%          \hline
%     \end{tabular}
    
% \end{table}
\section{Experiment Result and Discussion}

To evaluate the effectiveness of the proposed method, we compare it against baseline methods using three evaluation metrics: Attack Success Rate and Average Distance. These metrics allow us to quantify the success rate of our attacks and measure the magnitude of the introduced perturbations.

To demonstrate the impact of minimizing the Kullback-Leibler divergence ($D_{KL}$) on the logits, we conduct experiments to showcase how our approach effectively manipulates the logits to achieve the desired outcome. By minimizing $D_{KL}$ between the target logit distribution and the perturbed logit distribution, we aim to optimize the attack strategy and enhance the success rate.

Additionally, we investigated the influence of the regularization parameter on both the ASR and the Distance metrics. By varying the regularization parameter, we can analyze how it affects the success rate of the attack and the magnitude of the perturbations generated. This investigation provided insights into the role of regularization in balancing attack success and perturbation intensity.

\subsection{Effectiveness of Attack}
% \textbf{Effectiveness of Attack}

\begin{table}[!ht]
    \centering
    \caption{Attack Performance comparison}
    \renewcommand{\arraystretch}{1.5}
    \begin{tabular}{c|c|c}
        \hline
         \textbf{Method}  & \textbf{ASR}  & \shortstack{\textbf{Average Distance}}\\
         \hline
         FGSM       & $0.4517$            &  1.9843   \\
         BIM        & $0.4354$            &  1.1280   \\
         GM         & $0.4453$            &  1.2505   \\
         GM($\mathcal{L}^{2}$)     & $0.3736$            &  0.6312   \\
         SGM       & $0.3238$            &  0.5735   \\
        \hline
        \textbf{SWAP} & \textbf{0.5339} &\textbf{0.7661} \\
        \textbf{SWAP($\mathcal{L}^{2}$)} & \textbf{0.5231} & \textbf{0.5374}\\
         
%\multirow{2}{*}{\textbf{SWAP}} & \multirow{2}{*}{\shortstack{\textbf{0.5339}\\$\pm$\textbf{9.93e-4}}} & \multirow{2}{*}{\shortstack{\textbf{0.7661}\\ $\pm$\textbf{7.60e-2}}}  \\
%          & & \\
%\multirow{2}{*}{\textbf{SWAP($\mathcal{L}^{2}$)}} & \multirow{2}{*}{\shortstack{\textbf{0.5231}\\$\pm$\textbf{7.64e-4}}} & \multirow{2}{*}{\shortstack{\textbf{0.5374}\\$\pm$\textbf{3.74e-3}}}\\
 %         & & \\
         \hline
    \end{tabular}
   \label{table:comparison}
\end{table}

Table \ref{table:comparison} provides a comprehensive comparison of metrics for various algorithms, including SGM, GM($\mathcal{L}^{2}$), BIM, GM, FGSM, SWAP, and SWAP($\mathcal{L}^{2}$). Our proposed SWAP algorithm demonstrates remarkable success in both aspects: ASR and Average Distance. In specific, by comparing all baseline methods, SWAP and SWAP($\mathcal{L}^{2}$) achieved an ASR of over 50\%, surpassing all other methods. Furthermore, even with the inclusion of $\mathcal{L}^{2}$ regularization, the ASR performance remains unaffected, demonstrating the robustness of our approach. In terms of Average Distance, both SWAP and SWAP($\mathcal{L}^{2}$) outperform the other methods, achieving lower distances in successfully attacked cases. This indicates that our approach generates perturbations with a relatively lower magnitude compared to the other methods, resulting in less noticeable changes to the time series data. In sum, our proposed SWAP algorithm and its variant with $\mathcal{L}^{2}$ regularization consistently exhibit superior performance in terms of both ASR and Average Distance when compared to the baseline methods (SGM, GM($\mathcal{L}^{2}$), BIM, GM, and FGSM). These results highlight the effectiveness and efficiency of our approach in generating successful adversarial attacks on time series data.

\subsection{Performance Comparison}

Fig. \ref{fig:ASR comparison} shows the ASR comparison between SWAP,  GM, GM(\(\mathcal{L}^{2}\)), SGM, and the improved method SWAP(\(\mathcal{L}^{2}\)). As the goal is to maximize the ASR for achieving better attack, the points leaning to the left indicate that the dataset was attacked with higher ASR compared to our method. In other words, the closer a point is to the lower right corner, the more our method has improved performance. Based on Fig. \ref{fig:ASR comparison}, we can conclude that SWAP significantly outperforms other methods in terms of ASR. For all datasets, even in those where it is less effective, SWAP shows only subtle differences when compared to other methods, as evidenced by the points close to the diagonal.

As for the diagram of SWAP vs SWAP($\mathcal{L}^{2}$), the introduction of minor regularization($\mathcal{L}^{2}$) does not significantly reduce the reliability of our method, with almost all datasets dotted around the diagonal comparing with another inferior method. But it is worth mentioning that four datasets with low ASR since the introduction of regularization resulted in softer perturbance.

Fig. \ref{fig: Distance Comparison} shows the $\mathcal{L}^{2}$ distance metrics of SWAP($\mathcal{L}^{2}$) compared with other strategies. Compared to the FGSM, BIM and GM, our method shows more subtle perturbation, almost all datasets are located below the diagonal. As for the regularized GM with $\mathcal{L}^{2}$ and SGM, our algorithm is still competitive as we have more points that fall further off-diagonal to the lower right. The outstanding performance on average distance in terms of successful attacks, our algorithm shows stronger concealment and achieves a sneaky attack. Surprisingly, the regularized SWAP is slightly better than the unregularized SWAP in terms of ASR and the soft perturbance outperforms all implemented methods, which indicates that the regularization in our strategies can improve the anti-perception ability of noise without sacrificing ASR.

\subsection{Logits and Noise Analyzing}

Fig. \ref{fig:visualynalusis} illustrates the visual comparison among GM, SWAP, and their $\mathcal{L}^{2}$ strategy. The top four figures display the predicted and perturbed logits. The dissimilarity between the two logit distributions in GM-based methods is significantly larger than that in the SWAP-based method, as evident from both the $D_{KL}$ values and the visual logit plot. Our algorithm focuses on lowering the rank of the predicted logits to the second rank while preserving their values as much as possible while keeping the logits of other classes consistent. Fig. \ref{fig:visualynalusis} demonstrates that the $D_{KL}$ for the SWAP method is only 0.03, whereas it exceeds 3 for the GM-based methods. This high divergence is attributed to the numerous mismatched logits. In the same case of a successful attack, the GM-based methods introduced relatively high noise levels, with the original and $\mathcal{L}^{2}$ methods resulting in Euclidean distances of 1.36 and 0.42, respectively. In contrast, the SWAP methods exhibited negligible noise, with an Euclidean distance of 0.09. Furthermore, the noise level was significantly reduced to 0.03 when using the $\mathcal{L}^{2}$ regularized SWAP method.

From the last row of Fig. \ref{fig:visualynalusis}, it is evident that the GM method produces noticeable sawtooth shapes, with many crest points exceeding the boundary and resulting in noisy disturbances. Even with $\mathcal{L}^{2}$ regularization, the perturbations were still perceptible to the human eye, which poses a significant challenge for stealth attacks. In contrast, the perturbations generated by our SWAP strategy are faint and reliable. Interestingly, even slight regularization, with an $\alpha$ value as low as 0.01, is sufficient to achieve invisible and excellent attacks.

Furthermore, another noteworthy observation from Fig. \ref{fig:visualynalusis} is that our SWAP algorithm primarily focuses on changing the distribution between the top one and the top two logits. This approach ensures that the $D_{KL}$ value remains insignificant when the top two logits have remarkably close values, thus preserving the features of the original time series data to minimize noise. In contrast, the random selection of logits in the GM method demonstrates some detrimental effects. Without control over the logit distribution, the GM method tends to raise the confidence level of other logits that were originally zero. Consequently, the network assigns certain probabilities to these categories, indicating a certain degree of confidence in their corresponding features, resulting in excessive noise. Additionally, when encountering overconfidence in our predictions, the logits before the softmax activation can be considerably high. If the GM method accidentally selects logits with significantly lower confidence, it becomes extremely challenging for the deep neural model to promote the confidence of these logits, potentially leading to a failed attack. In contrast, our SWAP algorithm circumvents this issue by selectively manipulating the top two logits.

In summary, the SWAP strategy, which minimizes the $D_{KL}$ divergence, outperforms existing methods with a higher signal-to-noise ratio and a remarkable attack success rate.

\subsection{Parameter Tuning}

In the SWAP algorithm, the balance factor $\gamma$ was used to control the proportion between the first and second categories in their sum. By reducing the value of $\gamma$, the proportion of the largest logits decreased. When $\gamma$ approached 0, it indicated that we aimed to lower the desired confidence for the largest logits in the original distribution to near 0. Instead, we wanted to raise the value of the second-largest logits to 1. This manipulation of logits aimed to alter the model's prediction and achieve a successful adversarial attack.
\begin{figure}[h]
    \centering
    \includegraphics[scale=0.5]{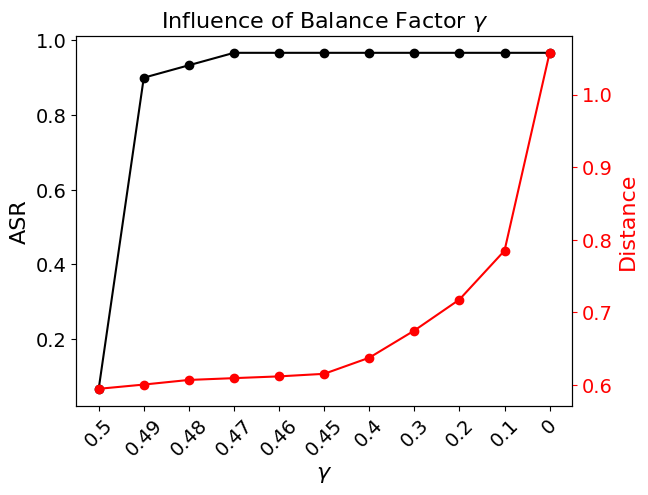}
    \caption{Scheme of adversarial attack, example data from Beef dataset, attacked by SWAP method}
    \label{fig:Incluence of gamma}
\end{figure}

From Fig. \ref{fig:Incluence of gamma}, we observed that the choice of $\gamma$ had a significant impact on the attack success rate (ASR). When $\gamma$ was set to 0.5, the ASR was extremely low because the prediction was near the decision boundary, making it difficult to alter the classification. However, as $\gamma$ decreased slightly below 0.5, the ASR increased rapidly to around 0.9. This was because reducing $\gamma$ pushed the logits of the incorrect class beyond the decision boundary, effectively changing the model's prediction. Therefore, increasing $\gamma$ beyond a certain threshold did not significantly improve the ASR and may disrupt the features of the original data without improving attack effectiveness. Hence, a suitable $\gamma$ value could be chosen within the range of 0.4 to 0.49 to maintain good consistency.

\begin{figure}[h]
    \centering
    \includegraphics[scale=0.5]{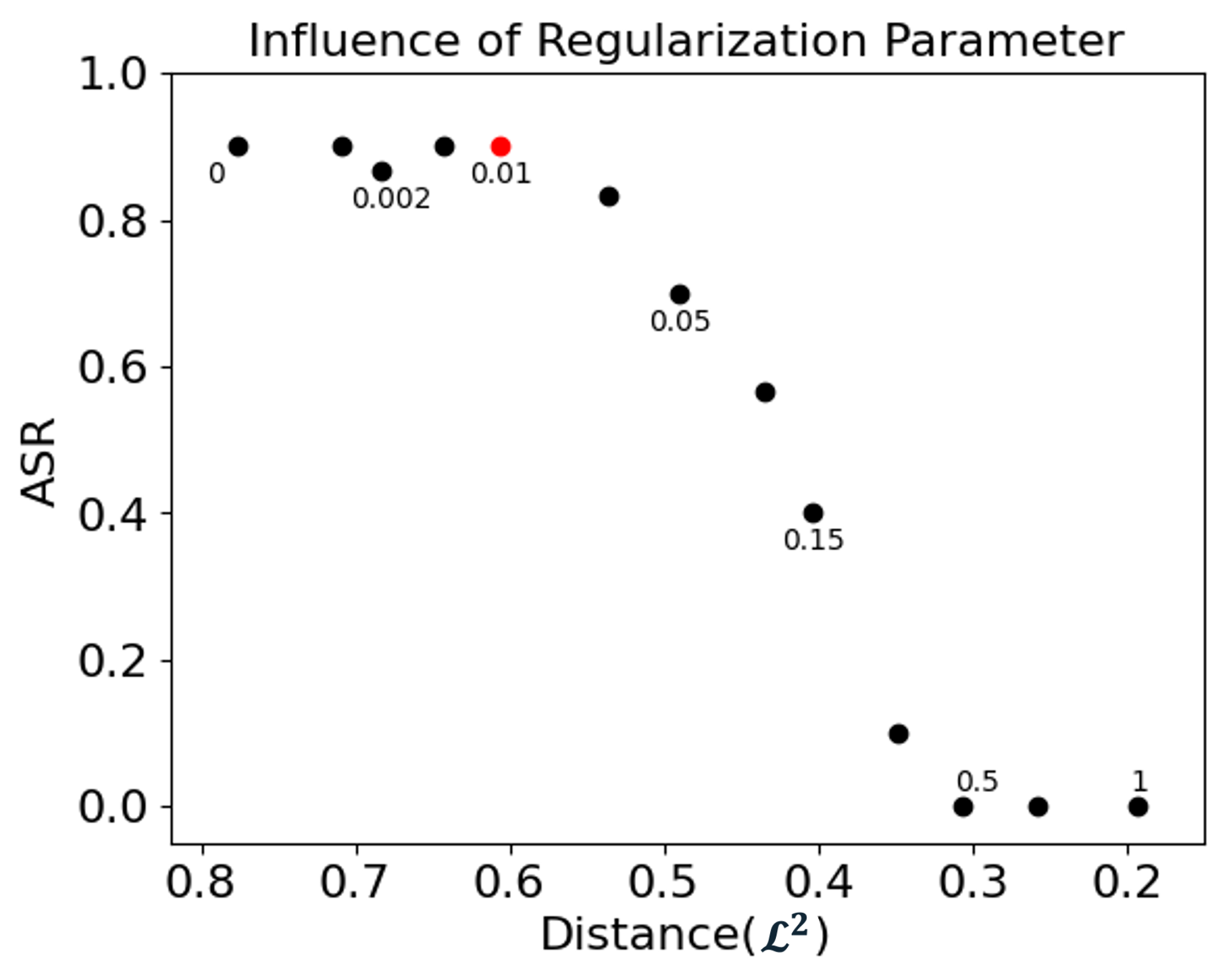}
    \caption{Scheme of adversarial attack, example data from Beef dataset, attacked by SWAP method}
    \label{fig:influence of regularization}
\end{figure}

Another important parameter in the SWAP algorithm was the coefficient of the regularization term, $\alpha$. Fig. \ref{fig:influence of regularization} demonstrated the trade-off between the distance and ASR for different $\alpha$ values. As $\alpha$ increases, both the distance and ASR decrease. However, when $\alpha$ exceeds 0.01, the ASR drops dramatically, indicating that the model became sensitive to the regularization term. Further reducing the distance through increased regularization does not offset the loss in ASR. Therefore, proper regularization, such as 1\% $\mathcal{L}^{2}$ regularization, can effectively reduce the distance and produce subtle perturbations while maintaining a high ASR. Interestingly, when $\alpha$ exceeded 0.5, the $\mathcal{L}^{2}$ norm regularization led to zero ASR, resulting in negligible noise influence.

\section{Conclusion}
In conclusion, The proposed adversarial attack framework, SWAP, presents a novel approach for targeting time series classification models. By leveraging the relationships between logits and the features of time series data, SWAP achieves targeted perturbations by strategically selecting the second-ranked logits and performing rank swapping with the prediction logits. Unlike existing methods, SWAP minimizes the manipulation of other logits, leading to more effective and stealthy attacks. Through extensive experiments, we have demonstrated the effectiveness of SWAP in generating adversarial samples. The results show that SWAP outperforms existing methods in terms of attack success rates while minimizing the amount of generated noise. 

\bibliographystyle{ieeetr}
\bibliography{reference}

\end{document}